\documentclass{ecai} 

\usepackage{latexsym}
\usepackage{amssymb}
\usepackage{amsmath}
\usepackage{amsthm}
\usepackage{booktabs}
\usepackage{enumitem}
\usepackage{graphicx}
\usepackage{color}
\usepackage{subfig}
\usepackage{mathtools}
\usepackage{algorithm}
\usepackage{algorithmic}

\usepackage[symbol]{footmisc}

\newtheorem{proposition}{Proposition}

\newtheorem{definition}{Definition}

\newcommand{\BibTeX}{B\kern-.05em{\sc i\kern-.025em b}\kern-.08em\TeX}

\DeclareMathOperator*{\argmin}{arg\,min}

\begin{document}


\begin{frontmatter}


\paperid{1985} 


\title{Offline Model-Based Reinforcement Learning with Anti-Exploration}


\author[1]{
\fnms{Padmanaba}~\snm{Srinivasan}\thanks{Corresponding Author. Email: ps3416@imperial.ac.uk}
}

\author[1]{
\fnms{William}~\snm{Knottenbelt}
}

\address[1]{Department of Computing, Imperial College London}


\begin{abstract}
Model-based reinforcement learning (MBRL) algorithms learn a dynamics model from collected data and apply it to generate synthetic trajectories to enable faster learning. This is an especially promising paradigm in offline reinforcement learning (RL) where data may be limited in quantity, in addition to being deficient in coverage and quality. Practical approaches to \textit{offline} MBRL usually rely on ensembles of dynamics models to prevent exploitation of any individual model and to extract uncertainty estimates that penalize values in states far from the dataset support. Uncertainty estimates from ensembles can vary greatly in scale, making it challenging to generalize hyperparameters well across even similar tasks. In this paper, we present \underline{Mo}rse \underline{Mo}del-based offline RL (MoMo), which extends the \textit{anti-exploration} paradigm found in offline model-free RL to the model-based space. We develop model-free and model-based variants of MoMo and show how the model-free version can be extended to detect and deal with out-of-distribution (OOD) states using explicit uncertainty estimation without the need for large ensembles. MoMo performs offline MBRL using an anti-exploration bonus to counteract value overestimation in combination with a policy constraint, as well as a truncation function to terminate synthetic rollouts that are excessively OOD. Experimentally, we find that both model-free and model-based MoMo perform well, and the latter outperforms prior model-based and model-free baselines on the majority of D4RL datasets tested.
\end{abstract}

\end{frontmatter}


\section{Introduction}

Reinforcement learning (RL) aims to learn policies for sequential decision-making that maximize an expected reward \citep{RN679}. In online RL, this means alternating between interacting with the environment to collect new data and then improving the policy using previously collected data. Model-based reinforcement learning (MBRL) denotes techniques in which an established environment model or an approximation of the environment is used to simulate policy rollouts without having to directly query the environment \citep{RN897,RN898,RN679,RN896}.

Offline RL tackles problems where the policy cannot interact with and explore the real environment but instead can only access a static dataset of trajectories. The offline dataset can be composed of mixed-quality trajectories with poor coverage of the state-action space. With no ability to extrapolate beyond the offline dataset, standard model-free offline RL algorithms often apply systematic pessimism by either injecting penalties into the estimation of values or by constraining the policy \citep{RN701,RN761} to remain in action support. Offline MBRL can offer more flexibility by augmenting the offline dataset with additional synthetic rollouts that expand coverage, while also remaining within the dataset support \citep{RN899,RN900,RN901}. 

MBRL is fraught with danger: approximation errors in the dynamics model can cause divergence in temporal difference updates \citep{RN902}, errors can accumulate over multistep rollouts, resulting in performance deterioration if unaddressed \citep{RN739}, and the dynamics model can be exploited by the agent \citep{RN903,RN904,RN905}. In offline MBRL, these issues are amplified due to the double dose of extrapolation error present in both dynamics and value functions. Consequently, standard online MBRL methods perform poorly when directly applied offline \citep{RN900}. 

Offline RL approaches typically fall into one of two categories: 1) \textit{critic regularization} methods which penalize OOD action-values or 2) \textit{policy constraint} methods that minimize a divergence between the learned policy and (a potentially estimated) behavior policy. Studying model-free offline RL literature exposes a variety of approaches in both paradigms \citep{RN701,RN761}, but offline MBRL methods exhibit far less diversity \citep{RN929}.

Many previous offline MBRL algorithms are based on MBPO \citep{RN739} and perform conservative value updates that identify the increased epistemic uncertainty in out-of-distribution (OOD) regions using dynamics ensembles \citep{RN905} or actively penalize all action-values from synthetic samples \citep{RN901}. Ultimately, these fall firmly under the critic regularization paradigm, and their empirical evaluation reveals that they perform little better than their model-free counterparts \citep{RN770,RN771}. A small number of methods implement policy-constrained offline MBRL \citep{RN915,RN916}, though these do not realize the benefits of augmenting training with samples from the dynamics model.

Our approach to offline MBRL is based on the \textit{anti-exploration} \citep{RN917} paradigm originating in model-free offline RL which uses a policy constraint in combination with an anti-exploration bonus to curb value overestimation. We train a Morse neural network \citep{RN870} and exploit its properties as an implicit behavior model, and an OOD detector to extend model-free anti-exploration for MBRL, which realizes the benefits of both policy-constrained offline RL \textbf{and} training with synthetic data. 

\noindent \textbf{Contributions}\quad In this work, we present \underline{Mo}rse \underline{Mo}del-based offline RL (MoMo) which performs offline model-free RL using anti-exploration. MoMo trains a dynamics model and a Morse neural network model, the latter of which is an energy-based model that learns an \textit{implicit} behavior policy and an OOD detector. We demonstrate how this can be incorporated into a model-free anti-exploration-based offline RL method and then adapt it to detect and truncate rollouts in the dynamics model that are becoming excessively out-of-distribution. The Morse neural network's uncertainty estimation helps us avoid the need for large dynamics ensembles -- we can instead use a single dynamics model. Furthermore, the Morse network's calibrated uncertainty estimates allow for better hyperparameter generalization across similar tasks.

Our experiments evaluate both model-free and model-based versions of MoMo on the D4RL Gym Locomotion and Adroit tasks \citep{RN731} and establish MoMo's performance to be equivalent to or better than recent baselines in 17 of 24 datasets. In-depth ablation experiments analyze MoMo's hyperparameter sensitivity and the effectiveness of anti-exploration value penalties in offline MBRL.

\section{Related Work}

\noindent \textbf{Reinforcement Learning}\quad We consider RL in the form of a Markov Decision Process (MDP), $\mathcal{M} = \{\mathcal{S}, \mathcal{A}, R, T, P_0, \gamma\}$, which consists of a tuple with state space $\mathcal{S}$, action space $\mathcal{A}$, a scalar reward $R(s, a)$, transition dynamics $T$, initial state distribution $P_0$ and discount factor $\gamma \in [0, 1)$ \citep{RN679}. RL aims to learn a policy $\pi$ that will execute an action $a = \pi(s)$ that maximizes the expected discounted reward ${J(\pi) = \mathbb{E}_{\tau \sim T_{\pi}(\tau)} \left[ \sum_{t=0}^{T} \gamma_t R(s_t, a_t) \right]}$ where ${T_{\pi}(\tau) = P_0 (s_0) \prod_{t=0}^{T} \pi(a_t \mid s_t) T(s_{t+1} \mid s_t, a_t)}$ is the trajectory under $\pi$. 

\noindent \textbf{Model-Free Offline RL}\quad Offline RL methods are designed to learn from a static dataset $\mathcal{D}$, which consists of interactions collected by applying a policy $\beta$, without access to the environment \citep{RN761,RN684,ijcai2024p545}. The offline dataset may provide poor coverage of the state-action space, and so when using function approximators, the policy must be constrained to select the actions within the support of the dataset. Offline RL methods (generally) belong to one of two families of approaches: 1) critic regularization or 2) policy constraint \citep{RN701,RN761}. Uncertainty-based model-free methods use large ensembles \citep{RN770,RN771} to perform a variation of critic regularization.

\noindent \textbf{Anti-Exploration}\quad \citet{RN917} pose model-free offline RL as an exercise in \textit{anti-exploration}, an approach in which the actor is constrained via an explicit divergence minimization term in the policy improvement step and policy evaluation is augmented with an anti-exploration bonus (value penalty) to counteract overestimation in off-policy evaluation. This bears some similarity to the BRAC framework \citep{RN699} which regularizes (either policy or value) using an explicit divergence between the current and estimated behavior policies. 

Several works propose different heuristics for estimating divergence in anti-exploration algorithms \citep{RN847,RN842}, though thus far anti-exploration has remained exclusive to model-free methods.

\noindent \textbf{Offline Model-Based RL}\quad Offline MBRL injects samples from a learned dynamics model into the RL learning process and leverages the diversity of simulated rollouts to augment the offline dataset with additional data. In online MBRL, \citet{RN739} introduce MBPO which uses an ensemble of dynamics models and performs Dyna-style policy learning \citep{RN897,RN898}. MBPO's ensemble does not discourage OOD exploration; \citet{RN900} address this by penalizing reward with ensemble uncertainty and \citet{RN899} use ensemble uncertainty to detect, penalize and terminate synthetic rollouts when they are excessively OOD. Several works develop methods to better estimate and penalize values using different uncertainty estimation techniques \citep{RN908,RN909}. 

COMBO \citep{RN901} performs CQL-like critic regularization \citep{RN910} on samples from model rollouts. This adversarial approach to offline MBRL is further developed by \citet{RN911}, \citet{RN912}, \citet{RN913} and \citet{RN914} who present alternative approaches to learning pessimistic value functions.

Policy constraint-based offline MBRL embraces an overarching goal to direct the policy towards states with known dynamics when starting in unknown ones. \citet{RN915} train ensemble dynamics and a behavioral transition model that is used to ensure that the next state the policy will transition to is within the dataset support. \citet{RN916} train an inverse dynamics model that constrains the policy to stay close to the behavior policy. Neither method performs policy rollouts using dynamics models, as estimates of the behavior policy do not extrapolate well. Their lack of training on synthetic data puts these methods at odds with critic-regularized offline MBRL.

\section{Preliminaries}

\citet{RN870} develop the Morse neural network, an uncertainty estimator that produces an unnormalized density $M (x) \in [0, 1]$ in an embedding space $\mathbb{R}^e$. At its modes, the density achieves a value of 1 and approaches 0 with increasing distance from the mode. A Morse neural network is defined by the combination of a neural network function approximator and a Morse kernel:
\begin{definition}[Morse Kernel]
    \label{def: morse kernel}
    A Morse Kernel is a positive definite kernel $K$ that is applied on a space $Z = \mathbb{R}^e$, $K(z_1, z_2)$ and takes its values in the interval $[0, 1]$ with $K(z_1, z_2) = 1$ only when \ $z_1 = z_2$.
\end{definition}

All kernels that use a measure of divergence are Morse kernels \citep{RN874}. A commonly used kernel is the radial basis function (RBF) kernel with scale parameter $\lambda > 0$
\begin{equation}
    \label{eq: rbf kernel}
    K(z, t) = e^{- \lambda \lvert\lvert z - t \rvert\rvert^2}.
\end{equation}

The Morse kernel describes how close an embedding $z$ is to the target $t$. The relationship between the input $x$ and its corresponding embedding $z$ is determined by a neural network $f_{\psi}: X \rightarrow Z$, $X \in \mathbb{R}^d$ and $Z \in \mathbb{R}^e$ with parameters $\psi$. Combining the Morse kernel with a neural network yields the Morse neural network:
\begin{definition}[Morse Neural Network]
    \label{def: morse neural network}
    A Morse neural network is comprised of a function $f_{\psi}: X \rightarrow Z$ and a Morse Kernel $K(z, t)$ where $t \subset Z$ is a target embedding where its size and value are chosen as a model hyperparameter. The Morse neural network is given by $M_{\psi} (x) = K(f_{\psi} (x), t)$.
\end{definition}

Using Definitions~\ref{def: morse kernel} and \ref{def: morse neural network}, if $x$ maps to a mode in the level set of the submanifold $Z$, then $M_{\psi}(x) = 1$. We interpret $M_{\psi}(x)$ as the \textit{certainty} that the sample $x$ is from the training dataset and $1 - M_{\psi}(x)$ is an estimate of the epistemic uncertainty of $x$. The function $-\log M_{\psi} (x)$ measures measures the distance between $z$ and the nearest mode at $t$ of the set of modes $M$:
\begin{equation}
    \label{eq: closest neighbor property}
    d(z) = \min_{t \in M}\quad d(z, t),
\end{equation}

The trained Morse neural network offers the following properties \citep{RN870}:
\begin{enumerate}
    \label{list: morse network properties}
    \item $M_{\psi}(x) \in [0, 1]$.
    \item $M_{\psi}(x) = 1$ at all mode submanifolds.
    \item $-\log M_{\psi} (x) \geq 0$ is a squared distance that satisfies the Morse--Bott non-degeneracy condition at the mode submanifolds \citep{RN872}. 
    \item $M_{\psi}(x)$ is an exponentiated squared distance, the function is distance aware in the sense that as $f_{\psi} (x) \rightarrow t, M_{\psi}(x) \rightarrow 1$.
\end{enumerate}

\noindent See \citet{RN870} for the proof of each property.

\section{Morse Model-Based Offline RL}

In this section, we describe our approach to policy-constrained offline MBRL that can use samples from synthetic rollouts to augment the offline dataset. We begin by adapting the Morse neural network for offline RL as a method for anti-exploration. Then in Section~\ref{subsec: extending to offline mbrl} we show how this can be extended to MBRL.

\subsection{Morse Neural Networks Learn an Implicit Behavior Policy}

\noindent \textbf{Morse Networks for Offline RL}\quad The offline RL dataset consists of $K$ tuples $\mathcal{D} = \{s, a, r, s'\}_{k = 1}^{K}$ that provide partial coverage of the state-action space. We train a Morse neural network to learn a conditional uncertainty model over actions in the dataset. Adapting the objective from \citet{RN870}, we minimize the KL divergence $D_{\text{KL}} (\mathcal{D} (s, a) \mid\mid M_{\psi} (s, a))$ using:
\begin{equation}
    \min_{\psi} \mathbb{E}_{s, a \sim \mathcal{D}} \left[ \log \frac{\mathcal{D} (s, a)}{M_{\psi} (s, a)} \right] + \int M_{\psi} (s, a) - \mathcal{D} (s, a)\ da,
\end{equation}

\noindent optimizing with respect to $\psi$, we minimize the empirical loss:
\begin{align}
    \label{eq: morse training objective}
    \nonumber \min_{\psi}\quad &\mathbb{E}_{s, a \sim \mathcal{D}} \left[ - \log K( f_{\psi} (s, a), t) \right] \\ +\ &\mathbb{E}_{\substack{s \sim \mathcal{D} \\ a_{\text{u}} \sim \mathcal{D}_{\text{uni}}}} \left[ K( f_{\psi} (s, a_{\text{u}}), t) \right],
\end{align}
\noindent where $\mathcal{D}_{\text{uni}}$ denotes a uniform distribution over the action space. We outline the Morse neural network training procedure in Algorithm~\ref{algo: morse neural network training}.

\begin{algorithm}[tb]
    \caption{Morse Neural Network Training}
    \label{algo: morse neural network training}
    \textbf{Input}: Offline dataset $\mathcal{D} = \{s, a, r, s'\}$\\
    \textbf{Initialize}: Initialize Morse network $M_{\psi}$. 
    \begin{algorithmic}[0] 
        \FOR{{t = 1} \TO {T}}
        \STATE $(s, a) \sim \mathcal{D} \quad \triangleright\ $ Sample real state-action tuples
        \STATE $a_{u} \sim \mathcal{D}_{\text{uni}} \quad \triangleright\ $ Sample random actions
        \STATE Update $\psi$ using Equation~\ref{eq: morse training objective}
        \ENDFOR
    \STATE \textbf{return} $M_{\psi}$
    \end{algorithmic}
\end{algorithm}

The Morse neural network is an energy-based model (EBM) \citep{RN884,RN870}:
\begin{proposition}
    \label{prop: morse network is an ebm}
    The Morse neural network is an EBM: ${E_{\psi}(x) = -\log M_{\psi(x)}}$.
\end{proposition}

See \citet{RN870} for the proof.

By interpreting the Morse neural network as an EBM we can recover a behavior policy density using the Boltzmann distribution:
\begin{align}
    \beta (a | s) &= \frac{M_{\psi}(s, a)}{Z_{\psi} (s)} 
    \\
    Z_{\psi} (s) &= \int_{\mathcal{A}} M_{\psi}(s, a)\ da,
\end{align}

\noindent where $Z_{\psi} (s)$ is a per-state normalization constant that, in practice, is intractable for continuous actions. This yields an implicit behavior policy:
\begin{align}
    \log \beta(a | s) &= \log M_{\psi} (s, a) - \log Z_{\psi} (s)
    \\
    &\leq \log M_{\psi} (s, a) \leq 0,
\end{align}
    
\noindent where the logarithm of the Morse neural network is an upper bound on the behavior density. Hence, maximizing certainty (or conversely, minimizing the distance $-\log M_{\psi}$ in the submanifold to a mode) will maximize log-likelihood under the implicit behavior policy. This also has the distinct advantage that all modes of the dataset occur with the same unnormalized density of one with no limiting inductive bias about the number of modes/policies that produced the training dataset. 

Many prior offline model-free methods estimate and constrain using explicit behavior policies \citep{RN684,RN699}. We perform similar constrained policy optimization with changes highlighted in \textcolor{blue}{blue}:
\begin{align}    
    \max_{\pi}\quad \mathbb{E}_{\substack{s \sim \mathcal{D} \\ a \sim \pi}} \left[ Q_{\theta}(s, a) - D_{KL}(\pi \mid\mid \beta) \right]
    \\
    \label{eq: policy objective}
    \approx \max_{\pi}\quad \mathbb{E}_{\substack{s \sim \mathcal{D} \\ a \sim \pi}} \left[ Q_{\theta}(s, a) + \textcolor{blue}{\log M_{\psi}(s, a)} \right].
\end{align}

This policy optimization is identical to that of prior anti-exploration methods \citep{RN723,RN842,RN847}, where optimizing $\log M_{\psi}$ allows minimization of the reverse KL divergence.

Following prior anti-exploration approaches, we use the following value function update with the anti-exploration bonus in \textcolor{red}{red}:
\begin{align}
    \label{eq: critic objective}
    \nonumber \min_{\theta}\quad \mathbb{E}_{\substack{s, a, s' \sim \mathcal{D} \\ a' \sim \pi}} \left[ (Q_{\theta} (s, a) - y)^2 \right]
    \\
    y = r + \gamma (Q_{\bar{\theta}} (s', a') + \textcolor{red}{\log M_{\psi} (s', a')}),
\end{align}

\noindent where $\theta$ denotes the parameters of the Q function and $\bar{\theta}$ the parameters of the target Q function.

Note that in Equation~\ref{eq: policy objective} and \ref{eq: critic objective} we disregard the scalar multiplier typically included with the $\log M_{\psi}$ term. We implicitly set this to one and instead use the kernel scale $\lambda$ in Equation~\ref{eq: rbf kernel} to control the behavioral cloning and anti-exploration tradeoff. 

\begin{figure*}[ht]
    \centering
    \subfloat[\centering Multimodal actions in a two-state dataset]{{\includegraphics[width=0.4\columnwidth]{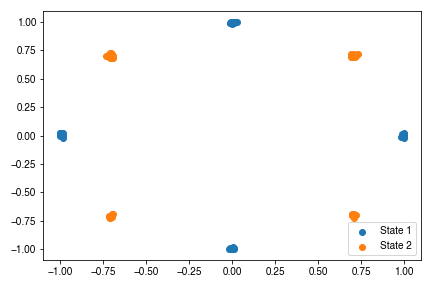} }}%
    \subfloat[\centering State 1, $\lambda = 0.1$]{{\includegraphics[width=0.4\columnwidth]{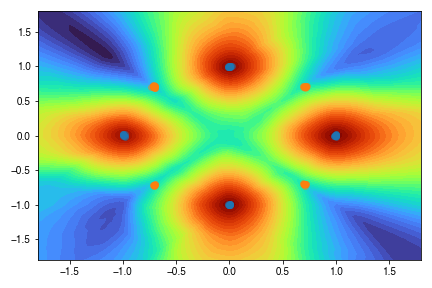} }}%
    \subfloat[\centering State 1, $\lambda = 1.0$]{{\includegraphics[width=0.4\columnwidth]{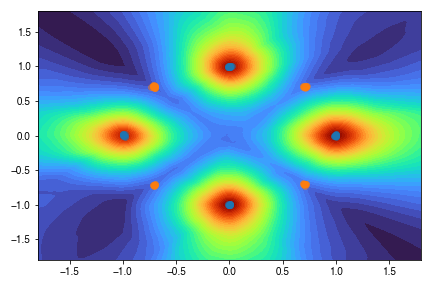} }}%
    \subfloat[\centering State 1, $\lambda = 2.0$]{{\includegraphics[width=0.4\columnwidth]{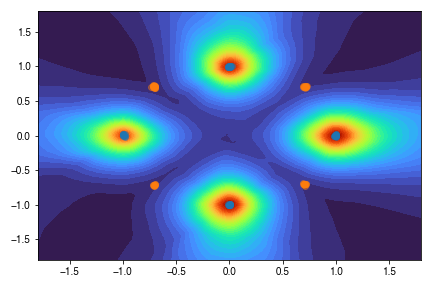} }}%
    \subfloat[\centering State 1, $\lambda = 4.0$]{{\includegraphics[width=0.4\columnwidth]{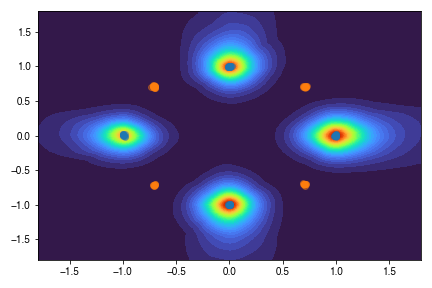} }}%
    \\
    \subfloat[\centering Density colormap]{{\includegraphics[width=0.4\columnwidth]{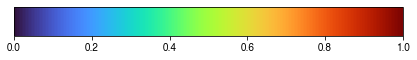} }}%
    \subfloat[\centering State 2, $\lambda = 0.1$]{{\includegraphics[width=0.4\columnwidth]{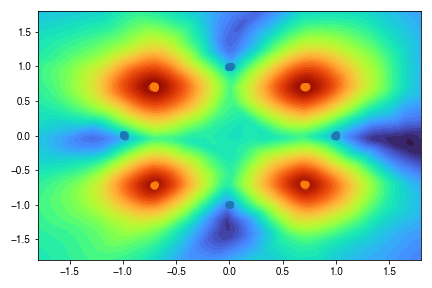} }}%
    \subfloat[\centering State 2, $\lambda = 1.0$]{{\includegraphics[width=0.4\columnwidth]{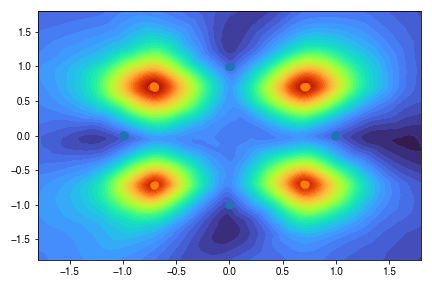} }}%
    \subfloat[\centering State 2, $\lambda = 2.0$]{{\includegraphics[width=0.4\columnwidth]{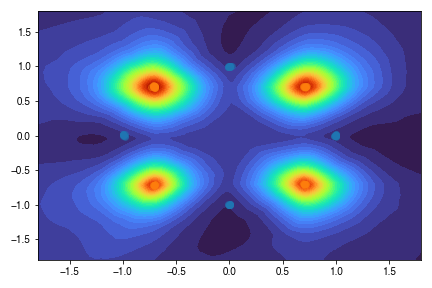} }}%
    \subfloat[\centering State 2, $\lambda = 4.0$]{{\includegraphics[width=0.4\columnwidth]{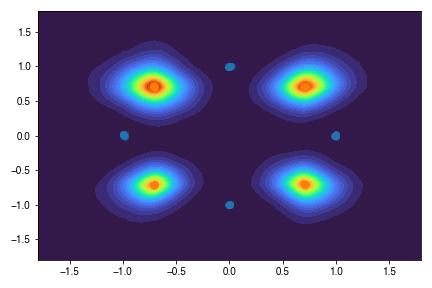} }}%
    \caption{Unnormalized density plots from a Morse neural network with RBF kernel (Equation~\ref{eq: rbf kernel}) for $\lambda = \{0.1, 1.0, 2.0, 4.0\}$. Eight action modes are uniformly spaced points on a unit circle with alternating modes assigned to each state. 64 action samples for each state are drawn with a standard deviation $0.01$ and clipped to lie in the range ${[-1.0, 1.0]}$. Subfig~{(a)} displays the training data and actions sampled for each state. Color-coded training points are overlaid on density plots for convenience. The Morse network (2 hidden layers of size 64, ReLU nonlinearity, gradient penalty, and layer normalization) captures all modes in each state. Note that the density plots cover a larger action range of ${[-1.8, 1.8]}$.}
    \label{fig: lambda illustrations}
\end{figure*}

\noindent \textbf{Control with Kernel Scale}\quad We illustrate how $\lambda$ can control the strength of the constraint with a didactic example in Figure~\ref{fig: lambda illustrations}. We select eight, two-dimensional action modes spaced uniformly on a unit circle that are assigned to one of two states in an alternating fashion. 64 action samples are drawn for actions in each state with a standard deviation of $0.01$. With a total of 128 samples from two states with four action modes each, we produce a synthetic dataset that exhibits a strong multimodal behavior policy (see {Figure~\ref{fig: lambda illustrations}(a))}. We fit a Morse neural network to the dataset using the objective in Equation~\ref{eq: morse training objective} and evaluate 16\,000 uniformly spaced samples over the action space to produce unnormalized densities for each state for different $\lambda$ in Figure~\ref{fig: lambda illustrations}. As $\lambda$ increases, the degree of ``closeness'' in the embedding space decreases resulting in a more sharply constraining density.

\noindent \textbf{Analysis}\quad We aim to constrain the policy to match the occupancy measure of the behavior policy. Assuming deterministic transition dynamics, we consider the normalized occupancy measure of the behavior policy using an EBM (see Proposition~\ref{prop: morse network is an ebm}):
\begin{align}
    (1 - \gamma)\rho_{\beta}(s, a) = \frac{\exp (-E_\psi(s, a))}{Z_{\psi}(s)}.
\end{align}

We minimize the KL divergence between occupancies and show that the policy constraint term in Equation~\ref{eq: policy objective} upper-bounds the divergence:
\begin{align}
    D_{\text{KL}} (\rho_{\pi} \mid\mid \rho_{\beta}) &= \sum_{s, a} \rho_{\pi} (s, a) \frac{\log \rho_{\pi} (s, a)}{\log \rho_{\beta} (s, a)}
    \\
    &\leq \mathbb{E}_{\pi} \left[ -\log M_{\psi} (s, a)  \right] - \mathcal{H}(\pi) + \text{const},
\end{align}

\noindent therefore, to minimize the KL divergence we choose to minimize the upper bound:
\begin{align}
    \argmin_{\pi} D_{\text{KL}} (\rho_{\pi} \mid\mid \rho_{\beta}) \rightarrow \argmin_{\pi} \mathbb{E}_{\pi} \left[ -\log M_{\psi} (s, a)  \right].
\end{align}

We provide the proof in the Supplementary Material. This shows that the behavioral constraint in Equation~\ref{eq: policy objective} is sufficient to ensure that the policy is restricted to following supported trajectories.

Furthermore, using Proposition~\ref{prop: morse network is an ebm}, theorems on performance from \citet{RN885} apply. Namely, Theorem~{2} suggests that any behavior policy with an arbitrary Lipschitz constant can approximated by an implicit model with a bounded Lipschitz constant with low model error.

\subsection{Challenges when Constraining Offline MBRL}
\label{subsec: challenges in constraining}
Na\"ively applying our policy constraint to states from model rollouts will result in poor performance. The Morse neural network as an implicit behavior policy poorly extrapolates to unseen states; prior work by \citet{RN919} and \citet{RN885} shows that implicit models formed by fully connected neural networks can only perform linear function extrapolation, yet assuming that the behavior policy is linear can be overly restrictive.

This is a critical shortcoming of EBMs when used for the policy constraint and anti-exploration bonus in unknown states, which can lead to incorrect modes being learned/penalized. This drawback extends to prior offline MBRL methods using a dynamics constraint \citep{RN915,RN916} who subsequently restrict themselves to training on small perturbations of dataset states rather than using dynamics rollouts. To use dynamics models for data augmentations, we must ensure that rollouts remain within the data support, both so that the Morse-based constraint is valid and to prevent the exploitation of learned dynamics. The ability to combine policy constraint with rollout OOD detection is (to the best of our knowledge) unique to MoMo and in the next section, we describe how the uncertainty detection properties of the Morse neural network enable detection of when states in synthetic trajectories have moved too far from the dataset support.

\subsection{Extending to Offline MBRL}
\label{subsec: extending to offline mbrl}

MBPO-based offline MBRL algorithms \citep{RN739} typically penalize OOD regions by directly using ensemble uncertainty estimates \citep{RN899,RN900} or assuming that all states produced by the dynamics models are OOD \citep{RN901,RN912,RN911,RN913,RN914}. MOReL \citep{RN899} terminates the rollouts when the estimated uncertainty increases beyond a threshold. We perform a similar rollout truncation when the policy is excessively OOD using the Morse neural network. 

Our rollout truncation refers to stopping the rollout without performing any subsequent bootstrapping or episode termination (i.e., we do not update truncated action-values during policy evaluation). This differs slightly from the treatment in online RL as in the offline domain, truncation is not an environment property, but a property that we impose on the empirical MDP.

Let $\beta(a | s)$ denote the behavior policy, $s_t$ denote the state at time $t$, $a_t$ denote the action selected by an agent, $P_0$ denote the initial state distribution and $\mathcal{T}(s_{t+1} \mid s_{t}, a_{t})$ denote the state transition. We assume that the environment transition is deterministic and aim to train an agent to minimize the (cross) entropy under the behavior policy over a trajectory $\tau_{0:T}$:
\begin{align}
    \mathcal{H} (\tau_{0:T}) &= -\sum_{t=0}^{T} \mathbb{E}_{\substack{a_t \sim {\pi(s_t)}}} \left[ \log \beta (a_t | s_t) \right] \mid \substack{s_0 \sim P_0 \\ s_{t+1} \sim \mathcal{T}(\cdot \mid s_t, a_t)}
    \\
    &\approx -\sum_{t=0}^{T} \mathbb{E}_{\substack{a_t \sim {\pi(s_t)}}} \left[ \log M_{\psi} (s_t, a_t) - \log Z_{\psi}(s_t) \right]
    \\
    &= -\sum_{t=0}^{T} \mathbb{E}_{\substack{a_t \sim {\pi(s_t)}}} \left[ \log M_{\psi} (s_t, a_t) \right] + \text{const}
    \\
    \label{eq: behavioral entropy minimization}
    &\geq -\sum_{t=0}^{T} \mathbb{E}_{\substack{a_t \sim {\pi(s_t)}}} \left[ \log M_{\psi} (s_t, a_t) \right].
\end{align}

Using an accurate learned dynamics model $\hat{\mathcal{T}}$, we can estimate a lower bound of trajectory entropy using $M_{\psi}$. By performing additional manipulation, we can instead extract a \textit{certainty} estimate over a trajectory:
\begin{align}
    \exp(-\mathcal{H} (\tau_{0:T})) &\leq \exp \left( \sum_{t=0}^{T} \mathbb{E}_{\pi} \left[ \log M_{\psi} (s_t, a_t) \right] \right) 
    \\
    &= \prod_{t=0}^{T} \mathbb{E}_{\pi} \left[ M_{\psi} (s_t, a_t) \right]
    \\
    &\coloneq P_{M_{\psi}}(\tau_{0:T}) \in [0, 1]
\end{align}

\noindent where the final line is the Morse-probability that the trajectory is behavior-consistent. This is an upper-bound estimate of the negative-exponentiated entropy. Therefore, a low entropy trajectory has a high trajectory-probability $P_{M_{\psi}}(\tau_{0:T})$. With the ability to estimate the probability of a rollout, we design a function to truncate a rollout when its probability falls below a threshold. 

\noindent \textbf{Rollout Truncation}\quad We design a detection mechanism to identify when a trajectory is excessively OOD and truncate the synthetic rollout.
\begin{definition}[OOD Truncation Function]
Given a rollout, at each timestep compute the Morse-probability $M_{\psi} (s_t, a_t)$ and the trajectory-probability ${P_{M_{\psi}}(\tau_{0:t}) = P_{M_{\psi}}(\tau_{0:t-1}) \times M_{\psi} (s_t, a_t)}$ and truncate the episode if the probability is below a threshold:
    \begin{align}
        \label{eq: truncation function}
        \texttt{trunc}(P_{M_{\psi}}(\tau_{0:t})) = 
            \begin{cases}
                \text{True} &\quad\text{if } P_{M_{\psi}}(\tau_{0:t}) < \epsilon_{\text{trunc}}, 
                \\
                \text{False} &\quad\text{otherwise.} \\ 
            \end{cases}
    \end{align}
Here, $\epsilon_{\text{trunc}} \in [0, 1]$ is a hyperparameter to be chosen that determines how much a trajectory can deviate from a behaviorally consistent one before the rollout is truncated.
\end{definition}

Selecting $\epsilon_{\text{trunc}} = 1$ will demand perfect reproduction of behavior policy trajectories and (in the absence of dynamics model errors) will be equivalent to model-free RL using only the offline dataset. Selecting a lower threshold allows the policy to explore the dynamics-estimated environment with increasingly low probability trajectories permitted as $\epsilon_{\text{trunc}} \rightarrow 0$. Algorithm~\ref{algo: rollout generation} describes how we employ the truncation function when generating rollouts.

\begin{algorithm}[tb]
    \caption{Truncated Synthetic Rollouts}
    \label{algo: rollout generation}
    \textbf{Input}: Initial state $s_0 \sim P_0$, Morse neural network $M_{\psi}$ and MLE trained dynamics $\hat{T}$\\
    \textbf{Initalize}: $P_{M_{\psi}}(\tau_{0:0}) = 1$, $\mathcal{D}_{\text{model}} = \{\}$
    \begin{algorithmic}[0] 
        \FOR{{t = 1} \TO {T}} 
        \STATE $a_t \sim \pi(a_t \mid s_t) \quad \triangleright\ $ Get action 
        \\
        \STATE Get $M_{\psi}(s_t, a_t)$ and compute $P_{M_{\psi}}(\tau_{0:t})\quad \triangleright\ $ Trajectory prob. 
        \\
        \STATE $\texttt{trunc}(P_{M_{\psi}}(\tau_{0:t})) \quad \triangleright\ $ Truncate rollout
        \\
        \STATE $s_{t+1}, r_{t} \sim \hat{\mathcal{T}}(\cdot \mid s_t, a_t) \quad \triangleright\ $ Step dynamics
        \\
        \STATE $\mathcal{D}_{\text{model}} \leftarrow \mathcal{D}_{\text{model}} \cup \{s_t, a_t, r_t, s_{t+1}\} \quad \triangleright\ $ Append to  buffer
        \ENDFOR
    \STATE \textbf{return} $\mathcal{D}_{\text{model}}$
    \end{algorithmic}
\end{algorithm}

\subsection{Practical Implementation}

We discuss aspects of MoMo's implementation here and provide additional details in the Supplementary Material. 

\noindent \textbf{Morse Net}\quad Training an implicit model with bounded Lipschitz constant also bounds the model error \citep{RN885}. When using neural network function approximators, this can be realized in several ways. We use a maximum-margin gradient penalty \citep{RN923} with Lipschitz constant set to 1 and add the gradient penalty loss as an auxiliary objective to Equation~\ref{eq: morse training objective} with equal weighting.

\noindent \textbf{Deep Architectures}\quad Rank collapse limits the expressivity of deep neural networks in offline RL \citep{RN927,RN928}. Morse neural networks benefit from deep architectures \citep{RN870,RN885} which can suffer from implicit underparametrization. This is a concern in MoMo as we directly use the Morse network as a loss function. \citet{RN924} present D2RL, a deep fully-connected architecture that uses skip connections which alleviate rank collapse at minimal extra computational cost. We adopt this architecture for the Morse neural network. In addition, we use layer normalization \citep{RN606} for the Morse and critic networks which improves stability.

\noindent \textbf{Actor--Critic}\quad MoMo can be used with any model-free RL algorithm. We use TD3 \citep{RN714} as the base actor--critic algorithm and retain standard TD3 hyperparameters. We use the policy objective specified in Equation~\ref{eq: policy objective} and also normalize the Q values for easier optimization.

\noindent \textbf{Dynamics Model}\quad A hallmark of most previous offline MBRL is the training and use of an ensemble (typically 4-7) of dynamics models to both prevent exploitation of individual models and for improved uncertainty estimation. The Morse network is an uncertainty estimator that we use directly to identify when a rollout goes OOD and for the anti-exploration bonus. This negates the need for ensemble-based uncertainty. Consequently, we train and use a \textbf{single} Gaussian dynamics model for all tasks.

\noindent \textbf{Hyperparameters}\quad Model-based MoMo is an offline MBRL algorithm that largely inherits the hyperparameters of MBPO. Furthermore, model-based MoMo adds two new parameters: 1) kernel scale $\lambda$ and 2) truncation threshold $\epsilon_{\text{trunc}}$. Prior critic-regularizing offline MBRL methods demand per-dataset tuning of hyperparameters to achieve reported performance. Such fine-grained tuning requirements can hamper the adoption of algorithms in real-world applications. In our experiments, we use a single set of hyperparameters in each domain to evaluate MoMo (i.e.\ one constant set of hyperparameters for \textbf{all} Locomotion tasks and a separate constant set for \textbf{all} Adroit tasks). Our ablation experiments demonstrate robust performance over a range of sensible hyperparameters. For primary results, we use $\lambda = 1.0$ and $\epsilon_{\text{trunc}} = 0.95$ for Locomotion tasks and $\lambda = 2.0$, $\epsilon_{\text{trunc}} = 0.98$ for Adroit.

\section{Experiments}

\begin{figure*}[h]
    \centering
    \includegraphics[width=0.75\textwidth]{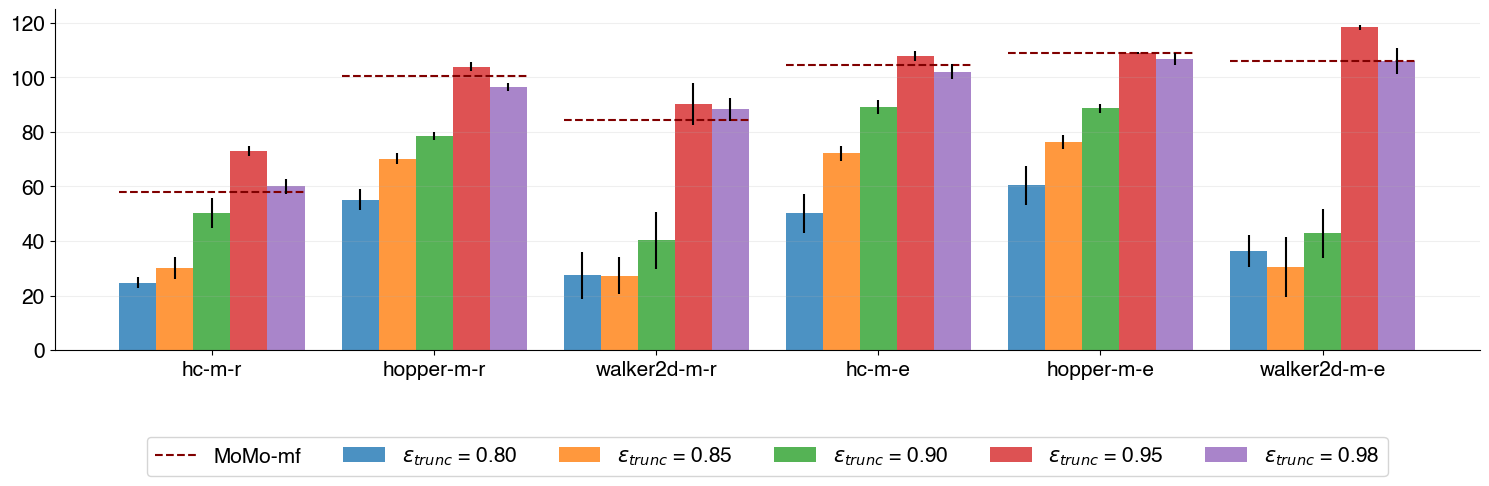}
    \caption{Ablations of $\epsilon_{\text{trunc}}$ on \texttt{-m-r} and \texttt{-m-e} datasets. The standard deviation over seeds is also included for each ablation. Note that $\epsilon_{\text{trunc}} = 0.95$ is the configuration used for MoMo-mb for primary results and MoMo-mf scores are included here for reference.}
    \label{fig: eps_thresh ablations}
\end{figure*}

\begin{figure*}[h]
    \centering
    \includegraphics[width=0.75\textwidth]{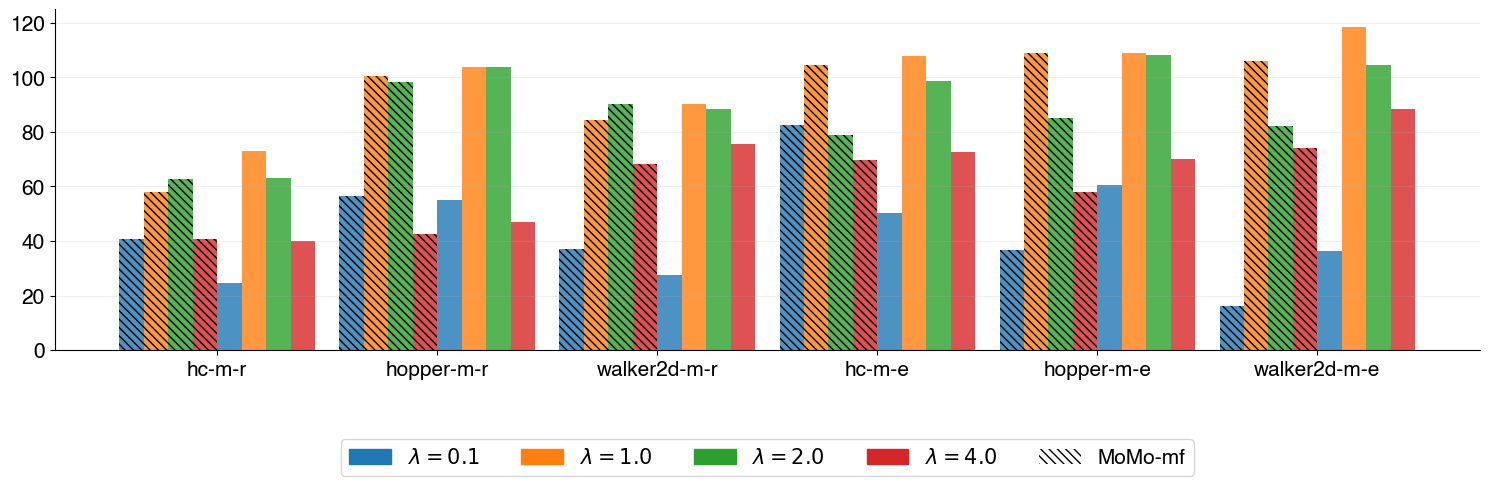}
    \caption{Ablations of $\lambda$ on \texttt{-m-r} and \texttt{-m-e} datasets. We use $\lambda = 1.0$ for primary results and perform additional sweeps over ${\lambda = [0.1, 2.0, 4.0]}$. The standard deviation over seeds is omitted for the sake of clarity. For MoMo-mb we use clear bars and maintain $\epsilon_{\text{trunc}} = 0.95$ in experiments.}
    \label{fig: lambda ablations}
\end{figure*}

We evaluate both model-based and model-free (i.e.\ generating no synthetic rollouts using learned dynamics) versions of MoMO and compare them with a range of baselines. We include the model-free baselines: CQL \citep{RN910}, IQL \citep{RN711}, TD3-BC \citep{RN719} and TD3-CVAE (model-free with anti-exploration) \citep{RN723}, the last of which is the progenitor of anti-exploration methods.

We also include the offline MBRL baselines: MoREL \citep{RN899}, MOPO \citep{RN900}, RAMBO \citep{RN911}, COMBO \citep{RN901}, ARMOR \citep{RN913}, MOBILE \citep{RN908} and OSR \citep{RN916}. The final three are recent methods that achieve state-of-the-art performance on benchmarks and are representative of the major techniques in modern offline MBRL.

ARMOR adopts an adversarial approach and performs value penalization on actions produced by the policy $\pi$ and increases values for actions produced by a reference policy $\pi_{\text{ref}}$, which must either be provided or be learned via behavioral cloning. MOBILE estimates uncertainty using estimated Q-values over the next states predicted by the dynamics models and penalizes the bootstrapped update with a penalty derived from the uncertainty. The authors of OSR propose two different methods of regularization: one that is policy constraining and the other a CQL-like critic regularization. OSR's policy is directed to recover towards in-dataset states when starting from unseen ones. We select results for the former version of OSR as a direct comparison to our policy constraint-based MoMo. 

In our experiments, we aim to answer the following questions:
\begin{itemize}
    \item How does MoMo perform in both model-based and model-free regimes?
    \item How important is the threshold $\epsilon_{\text{trunc}}$ for model-based performance?
    \item How sensitive is performance to the kernel scale $\lambda$?
    \item Is anti-exploration necessary for model-based MoMo? 
\end{itemize}

We answer the first question in the context of performance on D4RL Locomotion and Adroit datasets \citep{RN731} and compare scores with the aforementioned baselines. The remainder of the questions are answered via ablation experiments that analyze the impact of hyperparameters and design choices. All experiments were conducted over five seeds with evaluation over ten episodes.

\subsection{D4RL Performance}

We compare model-free and model-based MoMo with the aforementioned baselines for Gym Locomotion tasks in Table~\ref{tab: gym locomotion results} and Adroit tasks in Table~\ref{tab: adroit results}. 

\noindent \textbf{Gym Locomotion}\quad Gym Locomotion results show that model-based MoMo achieves the highest scores (either best or within 1 standard deviation of best) in 9 of the 12 datasets. Furthermore, MoMo-mb recovers expert performance ($\geq 95.0$) in 6 of 12 datasets with 100\% expert performance recovery on the \texttt{-m-e} datasets which contain a mixture of optimal and suboptimal trajectories. The \texttt{-r} datasets pose a significant challenge to all methods they contain highly suboptimal examples generated by a random uniform policy and all methods tend to perform poorly on these tasks. Comparing MoMo-mf to MoMo-mb, changing to the model-based domain improves performance by an average of 21.5\%, with marked improvements in the \texttt{-r}, \texttt{-m} and \texttt{-m-r} datasets. Finally, MoMo-mf outperforms both policy-constrained TD3-CVAE and dynamics-constrained OSR.

\noindent \textbf{Adroit}\quad The Adroit datasets pose a very different set of challenges to those in Locomotion tasks: Adroit datasets are marked by having few, human-generated examples (\texttt{-h}) or a mix between human and BC-trained behavior policies (\texttt{-c}). Consequently, learning algorithms must be able to deal with highly reward-disparate trajectories effectively and must generalize well. MoMo-mb achieves the best performance among all methods in 8 of 12 Adroit datasets and MoMo-mf consistently outperforms model-free baselines.

\begin{table*}[h]
    \caption{Normalized scores on D4RL Gym Locomotion datasets. Scores are taken from the reported score in the original work with top scores in \textbf{bold} and second-best \underline{underlined}. Model-free and model-based versions of MoMo are indicated by -mf and -mb, respectively, and scores include one standard deviation across seeds.} 
    \centering
    \resizebox{0.95\textwidth}{!}{
    \begin{tabular}{l|rrrr|rrrrrrr|rr}
        \toprule
         & \multicolumn{4}{|c|}{Model-free} & \multicolumn{6}{c|}{Model-based} & \multicolumn{1}{c|}{Inv-dyn} & \multicolumn{2}{c}{\textbf{Ours}} \\
        \midrule
        Dataset  & TD3-BC & CQL & IQL & TD3-CVAE & MoREL & MOPO & RAMBO & COMBO & ARMOR & MOBILE & OSR & MoMo-mf & MoMo-mb \\
        \midrule
        \texttt{hc-r} & 10.2 & 35.4 & - & 28.6  & 25.6 & 35.4 & \textbf{40.0} & 38.8 & - & 39.3 & 35.2 & 30.2 $\pm$ 2.3 & \underline{39.6 $\pm$ 3.7}
        \\
        \texttt{hopper-r} & 11.0 & 10.7 & - & 11.7 & \textbf{53.6} & 11.7 & 21.6 & 17.9 & - & \underline{31.9} & 10.3  & 9.4 $\pm$ 5.2 & 18.3 $\pm$ 2.8 
        \\
        \texttt{walker2d-r} & 1.4 & 2.7 & - & 5.5 & \textbf{37.3} & 13.6 & 11.5 & 7.0 & - & 17.9 & 13.5 & 20.7 $\pm$ 1.4 & \underline{26.8 $\pm$ 3.3}
        \\
        \texttt{hc-m} & 42.8 & 37.2 & 47.4 & 43.2 & 42.1 & 42.3 & \textbf{77.6} & 54.2 & 54.2 & 74.6 & 48.8 & 62.1 $\pm$ 1.4 & \underline{77.1 $\pm$ 0.9}
        \\
        \texttt{hopper-m} & 99.5 & 44.2 & 66.2 & 55.9 & 95.4 & 28.0 & 92.8 & 97.2 & 101.2 & \underline{106.6} & 95.2 & 95.7 $\pm$ 2.4 & \textbf{110.8 $\pm$ 2.3}
        \\
        \texttt{walker2d-m} & 79.7 & 57.5 & 78.3 & 68.2 & 77.8 & 17.8 & 86.9 & 81.9 & \underline{90.7} & 87.7 & 85.1 & 84.6 $\pm$ 2.8 & \textbf{95.0 $\pm$ 1.4}
        \\
        \texttt{hc-m-r} & 43.3 & 41.9 & 44.2 & 45.3 & 40.2 & 53.1 & 68.9 & 55.1 & 50.5 & \underline{71.7} & 46.8 & 58.1 $\pm$ 0.6 & \textbf{72.9 $\pm$ 1.8}
        \\
        \texttt{hopper-m-r} & 31.4 & 28.6 & 94.7 & 46.7 & 93.6 & 67.5 & 96.6 & 89.5 & 97.1 & 103.9 & 96.7 & \textbf{106.1 $\pm$ 1.0} & \underline{104.0 $\pm$ 1.8}
        \\
        \texttt{walker2d-m-r} & 25.2 & 15.8 & 73.8 & 15.4 & 49.8 & 39.0 & 85.0 & 56.0 & 85.6 & \underline{89.9} & 87.9 & 84.4 $\pm$ 2.9 & \textbf{90.4 $\pm$ 7.7}
        \\
        \texttt{hc-m-e} & 97.9 & 27.1 & 86.7 & 86.1 & 53.3 & 63.3 & 93.7 & 90.0 & 93.5 & \underline{108.2} & 94.7 & \textbf{110.7 $\pm$ 1.5} & 107.9 $\pm$ 1.9
        \\
        \texttt{hopper-m-e} & 112.2 & 111.4 & 91.5 & 111.6 & 108.7 & 111.9 & 83.3 & 111.1 & \underline{112.6} & 103.4 & \textbf{113.2} & 108.8 $\pm$ 0.6 & 109.1 $\pm$ 0.4
        \\
        \texttt{walker2d-m-e} & 101.1 & 68.1 & 109.6 & 84.9 & 95.6 & 44.6 & 68.3 & 103.3 & \underline{115.2} & 112.2 & 112.9 & 106.0 $\pm$ 8.3 & \textbf{118.4 $\pm$ 0.9}
        \\
        \bottomrule
    \end{tabular}
    }
    \label{tab: gym locomotion results}
\end{table*}

\begin{table}[h]
    \caption{Normalized scores on D4RL Adroit datasets. Baseline scores are taken from the original work with top scores in \textbf{bold} and second-best \underline{underlined}. Prior methods that do not evaluate on Adroit are excluded. Model-free and model-based versions of MoMo are indicated by -mf and -mb, respectively. We include the behavioral cloning baseline as it tends to perform well in Adroit tasks.}
    \centering
    \resizebox{0.98\columnwidth}{!}{
    \begin{tabular}{lrrrr|rr|rr}
        \toprule
        Dataset  & BC & CQL & IQL & TD3-CVAE & ARMOR & MOBILE & MoMo-mf & MoMo-mb \\
        \midrule
        \texttt{pen-h} & 34.4 & 37.5 & 71.5 & 59.2 & \underline{72.8} & 30.1 & 67.2 & \textbf{74.9}
        \\
        \texttt{hammer-h} & 1.5 & \textbf{4.4} & 1.4 & 0.2 & \underline{1.9} & 0.4 & 0.5 & 1.7
        \\
        \texttt{door-h} & 0.5 & \underline{9.9} & 4.3 & 0.0 & 6.3 & -0.2 & 6.8 & \textbf{11.3}
        \\
        \texttt{relocate-h} & 0.0 & 0.2 & 0.1 & 0.0 & \textbf{0.4} & - & 0.2 & \textbf{0.4}
        \\
        \texttt{pen-c} & 56.9 & 39.2 & 37.5 & 45.4 & 51.4 & 69.0 & \textbf{77.2} & \underline{74.1}
        \\
        \texttt{hammer-c} & 0.8 & \textbf{2.1} & \textbf{2.1} & 0.3 & 0.7 & 1.5 & 0.5 & 0.7
        \\
        \texttt{door-c} & -0.1 & 0.4 & 1.6 & 0.0 & -0.1 & \textbf{24.0} & 1.1 & \underline{5.8}
        \\
        \texttt{relocate-c} & \underline{-0.1} & \underline{-0.1} & -0.2 & -0.2 & \underline{-0.1} - & & -0.2 & \textbf{0.0}
        \\
        \texttt{pen-e} & 85.1 & 107.0 & - & \underline{112.3} & 112.2 & - & 97.4 & \textbf{132.6}
        \\
        \texttt{hammer-e} & 125.6 & 86.7 & - & \underline{128.9} & 118.8 & - & 126.9 & \textbf{131.8}
        \\
        \texttt{door-e} & 34.9 & \underline{101.5} & - & 59.4 & 98.7 & - & 76.7 & \textbf{103.4}
        \\
        \texttt{relocate-e} & 101.3 & 95.0 & - & \textbf{106.4} & 96.0 & - & 95.0 & \underline{104.7}
        \\
        \bottomrule
    \end{tabular}
    }
    \label{tab: adroit results}
\end{table}

\subsection{Ablating Threshold $\epsilon_{\text{trunc}}$}

The \texttt{-m-r} and \texttt{-m-e} Gym Locomotion tasks contain trajectories from two or more policies, which makes them candidate datasets for ablating MoMo hyperparameters. We evaluate performance when changing the truncation threshold parameter for values $\epsilon_{\text{trunc}} = \{ 0.80, 0.85, 0.90, 0.95, 0.98 \}$ in Figure~\ref{fig: eps_thresh ablations}. 

Across the board, reducing $\epsilon_{\text{trunc}}$ degrades performance. This is plausible, as reducing $\epsilon_{\text{trunc}}$ allows a trajectory to deviate more from one following the behavior policy before it is truncated. Overestimated action values inevitably propagate, resulting in an uncorrected distribution shift that results in low performance. Increasing $\epsilon_{\text{trunc}}$ to $0.98$ has a relatively small effect on performance. 

A higher $\epsilon_{\text{trunc}}$ increases the aggressiveness of truncation to ensure that trajectories are \textit{closer} to those following the behavior policy. In effect, this tries to ensure that the trajectories generated vary less from those in the dataset and consequently, we expect performance to be similar to model-free MoMo. In practice, dynamics model error may cause MoMo-mb to underperform compared to MoMo-mf.

\subsection{Sensitivity to $\lambda$}

We ablate the kernel scale parameter $\lambda$ on the \texttt{-m-r} and \texttt{-m-e} datasets for both MoMo-mf and MoMo-mb in Figure~\ref{fig: lambda ablations}. Increasing $\lambda$ \textit{sharpens} the density around the dataset modes (see Figure~\ref{fig: lambda illustrations}) and is equivalent to increasing the relative weighting of the policy constraint (recall that the log RBF kernel is $-\lambda \mid\mid z - t \mid\mid^2$). 

Figure~\ref{fig: lambda ablations} shows that using $\lambda = 0.1$ leads to a substantial performance drop in both model-free and model-based conditions as the constraint is too lax. Increasing $\lambda = 2.0$ has mixed results, and further increasing to $\lambda = 4.0$ likely overconstrains the policy. Interestingly, the change from $\lambda = 1.0 \rightarrow 2.0$ has a larger impact on MoMo-mf, on the \texttt{m-e} datasets where this change results in a {-22.9\%} performance change compared to a {-7.1\%} change for MoMo-mb.

\subsection{Importance of Anti-Exploration}

The primary mechanism by which MoMo encourages the policy to stay in-support is via a policy constraint. Previous work suggests that removing the policy constraint results in poor performance \citep{RN842}. The anti-exploration bonus (see Equation~\ref{eq: critic objective}) augments the bootstrapped return estimate, which in MoMo upper bounds the logarithm of the behavior policy density. This performs a regularized update that bounds KL regularized value iteration \citep{RN930,RN931} and curbs value overestimation caused by off-policy evaluation. We evaluate the impact of the anti-exploration bonus on MoMo-mb in Table~\ref{tab: anti-exploration ablations}. Removing the bonus causes performance to drop in almost all datasets, with an average performance decrease of 12.5\%, suggesting MoMo's distance-aware conservatism is critical to performance. The mixed impact on \texttt{-random} datasets may be due to the challenges of training the Morse network on actions sampled from a random policy.

\begin{table}[h]
    \caption{Normalized scores on D4RL datasets when removing the anti-exploration bonus. We also include the percentage change in performance.}
    \centering
    \resizebox{0.90\columnwidth}{!}{
    \begin{tabular}{lrr}
        \toprule
        Dataset  & MoMo-mb & \textbf{-}anti-exploration bonus \\
        \midrule
        \texttt{hc-r} & 39.6 $\pm$ 3.7 & \color{blue}{42.7 $\pm$ 3.2 (+7.8\%)}
        \\
        \texttt{hopper-r} & 18.3 $\pm$ 2.8 & \color{blue}{20.9 $\pm$ 4.5 (+14.2\%)}
        \\
        \texttt{walker2d-r} & 26.8 $\pm$ 3.3 & \color{red}{24.5 $\pm$ 1.7 (-8.6\%)}
        \\
        \texttt{hc-m} & 77.1 $\pm$ 0.9 & \color{red}{70.5 $\pm$ 1.1 (-8.6\%)}
        \\
        \texttt{hopper-m} & 110.8 $\pm$ 2.3 & \color{red}{92.4 $\pm$ 2.6 (-16.6\%)}
        \\
        \texttt{walker2d-m} & 95.0 $\pm$ 1.4 & \color{red}{74.9 $\pm$ 3.9 (-21.6\%)}
        \\
        \texttt{hc-m-r} & 72.9 $\pm$ 1.8 & \color{red}{66.1 $\pm$ 1.4 (-9.3\%)}
        \\
        \texttt{hopper-m-r} & 104.0 $\pm$ 1.8 & \color{red}{68.1 $\pm$ 1.9 (-34.5\%)}
        \\
        \texttt{walker2d-m-r} & 90.4 $\pm$ 7.7 & \color{red}{87.2 $\pm$ 5.8 (-3.5\%)}
        \\
        \texttt{hc-m-e} & 107.9 $\pm$ 1.9 & \color{red}{100.7 $\pm$ 3.1 (-6.7\%)}
        \\
        \texttt{hopper-m-e} & 109.1 $\pm$ 0.4 & \color{red}{102.6 $\pm$ 2.5 (-6.3\%)}
        \\
        \texttt{walker2d-m-e} & 118.4 $\pm$ 0.9 & \color{red}{98.2 $\pm$ 0.7 (-17.1\%)}
        \\
        \midrule
        Total & 970.3 & \color{red}{848.8  (-12.5\%)}
        \\
        \bottomrule
    \end{tabular}
    }
    \label{tab: anti-exploration ablations}
\end{table}

\section{Discussion}

\noindent \textbf{Computational Cost}\quad Model-based MoMo trains a dynamics model as well as the Morse neural network. This translates into an increase in total training time (pretraining dynamics and Morse network for 100\,000 steps followed by policy/value learning for 1 million steps) of approximately 65\% over model-free TD3-BC and 25\% over MBPO.

\noindent \textbf{Uncertainty Estimation}\quad Morse neural networks are a specific method of uncertainty estimation that also enjoy equivalency to EBMs, resulting in specific properties used in this work. Alternative methods of uncertainty estimation can be used, the Morse network itself being identical to other estimators depending on instantiation (see \citet{RN870} for a more detailed discussion). 

\noindent \textbf{Limitations}\quad The primary limitation of MoMo is the inability of the Morse network is extrapolate beyond the offline dataset state support, thus requiring the truncation function when using model-based MoMo, which introduces the $\epsilon_{\text{trunc}}$ hyperparameter. An alternative is to update the Morse network periodically using synthetically generated trajectories, though this will add the optimization challenges of adversarial training dynamics and increase training cost.

\section{Conclusion}

In this paper, we studied how a model-free offline RL paradigm could be extended to the model-based domain. We incorporated Morse neural networks into the anti-exploration framework to yield MoMo, which constrains both the policy and penalizes off-policy evaluation with an anti-exploration bonus. To overcome the limitations of model-free MoMo, we deploy Morse networks as both an EBM and an uncertainty estimator to truncate OOD trajectories. Experiments comparing both model-free and model-based versions of MoMo with recent baselines demonstrate MoMo's superior results as well as the benefits of incorporating dynamics models.

Ensemble-based uncertainty estimates can vary greatly in scale which necessitates per-dataset tuning to achieve reported performance. In contrast, MoMo uses identical hyperparameter values across tasks in the same domain, and subsequent ablation experiments demonstrate that performance is stable across a range of hyperparameters. Future work should analyze the approaches employed by modern model-free offline RL and explore how they may be adapted for offline MBRL.

\bibliography{mybib}

\end{document}



\begin{frontmatter}


\paperid{1985} 


\title{Offline Model-Based Reinforcement Learning with Anti-Exploration}


\author{\fnms{Padmanaba}~\snm{Srinivasan}\thanks{Corresponding Author. Email: ps3416@imperial.ac.uk}}

\author{\fnms{William}~\snm{Knottenbelt}\thanks{Email: wjk@imperial.ac.uk}}

\address{Department of Computing, Imperial College London}

\end{frontmatter}


\section{Proofs \& Derivations}

We make the following assumption.

\begin{assumption}
    \label{ass: determinsitic transitions}
    The environment transition is deterministic.
\end{assumption}

We reiterate the following proposition.

\begin{proposition}
    \label{prop: morse network is an ebm}
    The Morse neural network is an EBM \citep{RN884,RN870}: ${E_{\psi}(x) = -\log M_{\psi(x)}}$.
\end{proposition}

Proposition~\ref{prop: morse network is an ebm} allows us to estimate the behavior policy using the Boltzmann distribution:
\begin{align}
    \beta (a | s) &= \frac{M_{\psi}(s, a)}{Z_{\psi} (s)} 
    \\
    Z_{\psi} (s) &= \int_{\mathcal{A}} M_{\psi}(s, a)\ da,
\end{align}

thus, the Morse neural network is an \textit{implicit} behavioral policy. 

We also use the following lemma from \citet{RN676}({Lemma~3.1})
\begin{lemma}
    Let $\bar{\mathcal{H}} = -\sum_{s, a} \rho(s, a) \log \frac{\rho(s, a)}{\sum_{a'} \rho(s, a)}$. Then $\bar{\mathcal{H}}$ is strictly concave. For all $\pi \in \Pi$ and $\rho \in \mathcal{D}$, we have $\mathcal{H} (\pi) = \bar{\mathcal{H}} (\rho_{\pi})$ and $\bar{\mathcal{H}} (\rho) = \mathcal{H} (\pi_{\rho})$. 
\end{lemma}

Using Assumption~\ref{ass: determinsitic transitions} and Proposition~\ref{prop: morse network is an ebm}, we consider the normalized occupancy measure of the implicit behavior policy:
\begin{align}
    (1 - \gamma)\rho_{\beta}(s, a) = \frac{\exp (-E_\psi(s, a))}{Z_{\psi}(s)} = \frac{M_{\psi} (s, a)}{Z_{\psi} (s)}.
\end{align}

To effectively constrain the policy to select in-support actions, we minimize the KL divergence between occupancies and show that this results in our policy constraint:
\begin{align}
    D_{\text{KL}} (\rho_{\pi} \mid\mid \rho_{\beta}) &= \sum_{s, a} \rho_{\pi} (s, a) \frac{\log \rho_{\pi} (s, a)}{\log \rho_{\beta} (s, a)}
    \\
    &= \sum_{s, a} \rho_{\pi} \left( \log \rho_{\pi} - \log \rho_{\pi_{\beta}} \right)
    \\
    &= \sum_{s, a} \rho_{\pi} \left( \log \rho_{\pi} - \log \frac{M_{\psi}}{(1 - \gamma) Z_{\psi}} \right)
    \\
    &= \sum_{s, a} \rho_{\pi} \left( \log \rho_{\pi} - \log M_{\psi} + \log (1 - \gamma) Z_{\psi} \right)
    \\
    &= \mathbb{E}_{\pi} \left[ -\log M_{\psi} \right] + \sum_{s, a} \rho_{\pi} \log \rho_{\pi} + \text{const}
    \\
    \nonumber &= \mathbb{E}_{\pi} \left[ -\log M_{\psi} \right] 
        \\
        \nonumber &\quad+ \sum_{s, a} \rho_{\pi} \left( \log \rho_{\pi} + \log \sum_{a'} \rho_{\pi} - \log \sum_{a'} \rho_{\pi} \right) 
        \\ 
        &\quad+ \text{const}
    \\
    \nonumber &= \mathbb{E}_{\pi} \left[ -\log M_{\psi} \right] 
        \\
        \nonumber &\quad+ \sum_{s, a} \rho_{\pi} \log \frac{\rho_{\pi(s, a)}}{\sum_{a'} \rho_{\pi} (s, a')} 
        \\ 
         &\quad+ \sum_{s} \rho_{\pi} (s) \log \rho_{\pi} (s)+ \text{const}
    \\ 
    &= \mathbb{E}_{\pi} \left[ -\log M_{\psi} \right] - \bar{\mathcal{H}}(\rho_{\pi}) - \mathcal{H}(\rho_{\pi} (s)) + \text{const}
    \\
    &\leq \mathbb{E}_{\pi} \left[ -\log M_{\psi} \right] - \mathcal{H}(\pi) + \text{const}.
\end{align}

For a deterministic policy, we choose to minimize $-\log M_{\psi}$ which is an upper bound over the KL divergence:
\begin{align}
    \argmin_{\pi} D_{\text{KL}} (\rho_{\pi} \mid\mid \rho_{\beta}) \rightarrow \argmin_{\pi} \mathbb{E}_{\pi} \left[ -\log M_{\psi} (s, a)  \right].
\end{align}

When training a stochastic policy (Gaussian) \citep{RN712}, one should include the entropy maximization term to yield:
\begin{align}
    \argmin_{\pi} D_{\text{KL}} (\rho_{\pi} \mid\mid \rho_{\beta}) \rightarrow \argmin_{\pi} \mathbb{E}_{\pi} \left[ -\log M_{\psi} (s, a)  \right] - \mathcal{H} (\pi).
\end{align}


\section{Implementation Details}

Our implementation of MoMo is written in PyTorch \citep{RN895} and based on the offline RL algorithm implementations by \citet{RN820}. 

\noindent \textbf{Morse Neural Network}\quad We use the deep, D2RL architecture \citep{RN924} for the Morse neural network as it is resistant to rank collapse \citep{RN927,RN928}, with layer normalization \citep{RN606}. A behavior policy with arbitrary Lipschitz constant can be approximated by a Lipschitz-bounded implicit model \citep{RN885}; we use a maximum-margin gradient penalty \citep{RN923} with constant $L = 1$:
\begin{align}
    \nonumber \mathcal{L}_{\text{pen}} =  \sum_{i=1}^{K} \text{max} &\left( 0, \lvert\lvert f_{\psi} (s, a) \rvert\rvert_2 - L \right)^2
    \\
    + \text{max} &\left( 0, \lvert\lvert f_{\psi} (s, a_{u}) \rvert\rvert_2 - L \right)^2,
\end{align}

\noindent over $K$ real and fake training samples, and the penalty is weighted equally with the Morse training objective. We summarize Morse neural network hyperparameters in Table~\ref{tab: morse net hyperparameters}

\begin{table}
    \caption{Morse Neural Network hyperparameters.}
    \centering
    \begin{tabular}{lr}
        \toprule
        Hyperparameter & Value \\
        Hidden layer size & 256 \\
        Layer normalization & Yes \\
        Nonlinearity & ReLU \\
        Training batch size & 512 \\
        Optimizer & Adam \cite{RN484} \\
        Learning rate & 0.0003 \\
        Weight decay & 0.0 \\
        Training steps & 100\,000\\
        Kernel & RBF \\
        \bottomrule
    \end{tabular}
    \label{tab: morse net hyperparameters}
\end{table}

\noindent \textbf{Dynamics Model}\quad We train a single dynamics model for each dataset similarly to \citet{RN900}. We use a deeper neural network with five hidden layers and layer normalization. We summarize dynamics model hyperparameters in Table~\ref{tab: dynamics hyperparameters}

\begin{table}
    \caption{Dynamics Model hyperparameters.}
    \centering
    \begin{tabular}{lr}
        \toprule
        Hyperparameter & Value \\
        Hidden layer size & 256 \\
        Layer normalization & Yes \\
        Nonlinearity & ReLU \\
        Training batch size & 512 \\
        Optimizer & Adam \\
        Learning rate & 0.0003 \\
        Weight decay & 0.0 \\
        Training steps & 100\,000\\
        \bottomrule
    \end{tabular}
    \label{tab: dynamics hyperparameters}
\end{table}

\noindent \textbf{TD3}\quad We use the standard TD3 hyperparameters, with the addition of critic layer normalization which can improve stability \citep{RN892,RN807}. Hyperparameters for Model-based MoMo are in Table~\ref{tab: momo hyperparameters}.

\begin{table}
    \centering
    \caption{MoMo hyperparameters.}
    \begin{tabular}{lr}
        \toprule
        Hyperparameter & Value \\
        Hidden layer size & 256 \\
        Number of hidden layers & 2 \\
        Nonlinearity & ReLU \\
        Training batch size & 512 \\
        Optimizer & Adam \\
        Learning rate & 0.0003 \\
        Weight decay & 0.0 \\
        Training steps & $10^6$ \\
        Actor layernorm & No \\
        Critic layernorm & Yes \\
        $m$ (policy update frequency) & 2 \\
        Policy noise & 0.5 \\
        Noise clip & [-0.5, 0.5] \\
        Target update rate ($\rho$) & 0.005 \\
        Discount factor & 0.99 \\
        Exploration noise & 0.1 \\
        $\lambda$ & 1.0 (Locomotion), 2.0 (Adroit) \\
        $\epsilon_{\text{trunc}}$ (-mb) & 0.95 (Locomotion), 0.98 (Adroit) \\
        Rollout horizon (-mb) & 100 \\
        Real-to-synthetic ratio (-mb) & 0.5 \\
        \bottomrule
    \end{tabular}
    \label{tab: momo hyperparameters}
\end{table}



\bibliography{mybib}